\documentclass{article}

\usepackage[preprint]{neurips_2024}

\usepackage[utf8]{inputenc} 
\usepackage[T1]{fontenc}    
\usepackage{hyperref}       
\usepackage{url}            
\usepackage{booktabs}       
\usepackage{amsfonts}       
\usepackage{amsmath}
\usepackage{nicefrac}       
\usepackage{microtype}      
\usepackage{xcolor}         
\usepackage{multirow}
\usepackage{graphicx}
\usepackage{tabularx}
\usepackage{enumitem}
\usepackage{caption}

\title{MobileExperts: A Dynamic Tool-Enabled Agent Team in Mobile Devices}

\author{%
\textbf{Jiayi Zhang}$^{1,2}$,
\textbf{Chuang Zhao}$^3$ ,  
\textbf{Yihan Zhao}$^2$,
\textbf{Zhaoyang Yu}$^2$ ,\\
\textbf{Ming He}$^1$  \thanks{Ming He (E-mail: heming01@foxmail.com),
  is the corresponding author.} ,
\textbf{Jianping Fan}$^1$
   \vspace{.5em} 
  \\
  $^1$AI Lab at Lenovo Research, 
  $^2$Renmin University of China, \\
  $^3$The Hong Kong University of Science and Technology\\
}

\begin{document}

\maketitle
\begin{abstract}

The attainment of autonomous operations in mobile computing devices has consistently been a goal of human pursuit. With the development of Large Language Models (LLMs) and Visual Language Models (VLMs), this aspiration is progressively turning into reality. While contemporary research has explored automation of simple tasks on mobile devices via VLMs, there remains significant room for improvement in handling complex tasks and reducing high reasoning costs. In this paper, we introduce MobileExperts, which for the first time introduces tool formulation and multi-agent collaboration to address the aforementioned challenges. More specifically, MobileExperts dynamically assembles teams based on the alignment of  agent portraits with the human requirements. Following this, each agent embarks on an independent exploration phase, formulating its tools to evolve into an expert. Lastly, we develop a dual-layer planning mechanism to establish coordinate collaboration among experts. To validate our effectiveness, we design a new benchmark of hierarchical intelligence levels, offering insights into algorithm's capability to address tasks across a spectrum of complexity. Experimental results demonstrate that MobileExperts performs better on all intelligence levels and achieves $\sim$22\% reduction in reasoning costs, thus verifying the superiority of our design. 
\end{abstract}

\section{Introduction}

The human pursuit of creating autonomous device operation agents to perform tasks independently continues, intensified by the widespread adoption of personal electronic devices\cite{zhang2023appagent,wang2024mobileagent,wu2024oscopilot,zhang2024ufo,Zhao2022Winning,tan2024towards}. People leverage these intelligent assistants to help themselves, not only completing mundane tasks on a daily basis, but also surpassing their own capabilities to perform higher-level operations. In previous research, achieving autonomous planning and execution abilities in agents has been challenging due to limitations in intelligence levels. However, the advent and evolution of large language models (LLMs) bring forward the potential for its realization. A significant number of studies concentrate on LLM-based agents \cite{wang2024survey,guo2024large}and demonstrate remarkable competency in varied domains.

Recent researchers attempt to implement this concept using LLMs or VLMs. With the integration of a LLM/VLM core, it's possible to configure the agent's action space to interact with a device enabling the agent to operate and control the specific device autonomously. We refer to this type of work as "Device Operation Agent (DOA)". Among all possible device types, mobile devices, due to their close ties with human daily life, are the platform most urgently needing DOAs. Currently, research on mobile device DOA has become a popular field of focus.  For example, research work such as \cite{wen2024autodroid, zhang2023appagent, wang2024mobileagent} demonstrate the application of LLM and VLM to accomplish automated operational tasks.


However, based on the intelligence levels defined by \cite{li2024personal}, even the best-performing DOA is currently only at the level of "Simple Step Following" and "Deterministic Task Automation". Additionally, current UI-based DOAs are facing considerable challenges regarding reasoning efficiency and time cost. A typical UI operation may require 5 to 10 steps, while each step of an existing UI-based agent requires the invocation of the VLM model to choose the action, undoubtedly adding significant reasoning and time costs. 


To solve these problems, we introduce MobileExperts, a multimodal multi-agent framework for mobile devices on android, designed to enhance DOA intelligence and reduce reasoning costs and time. It features two key features:
\textit{\textbf{(1) Code-Combined Procedure Memory (Tool) Formulation: }}
To minimize reliance on the VLM model, we propose a new form of tool pattern: with the code writing ability of LLM, MobileExperts combines the basic actions of experts through code combination to form reusable code block tools. These tools are stored in the procedure memory shared by all experts in the system, thereby reducing the cost of tool formation and effectively improving operational efficiency. \textit{\textbf{(2) Expert Collaboration via Double-layer Planning: }}MobileExperts adopts double layer planning method to solve long-term planning problems: the first layer is the team task distributed layer, at this level, tasks are decomposed into dependent subtasks for experts to execute, these dependencies form a collaborative network between experts; the second layer is expert task decomposition layer, the subtasks received by the experts are further decomposed into smaller action units to gradually achieve the goal. By these methods, MobileExperts not only improves the intelligence level but also reduces the cost of reasoning and time, providing a more efficient task execution mode for the DOA field.

In addition, to effectively measure the capability of the MobileExperts framework, we proposed Expert-Eval to measure the performance of DOA at different intelligence levels. We compared MobileExperts with multiple baselines on this benchmark and conducted ablation experiments on the module to verify the effectiveness of the framework. We summarize the contributions of this paper as below:


\begin{itemize}[leftmargin=2em]
\setlength{\itemsep}{1pt}
\setlength{\parsep}{1pt}
\setlength{\parskip}{1pt}
    \item MobileExperts is an autonomous multimodal agent framework that excels in handling complex device operation tasks, which enhanced by dynamic agent collaboration mechanism.
    \item We propose an automatic code based tool formation method based on VLM, utilizing action trajectories accumulated within the interaction with environment, which fills the gap in tool formation for the mobile device operation field.
    \item We design a tiered intelligence evaluation benchmark, Expert Eval, to assess the performance of mobile device operation agents across tasks of varying complexities. 
\end{itemize}

\section{Related Work}
\subsection{LLM Based MultiAgent System}

Due to the extraordinary performance demonstrated by LLMs under the Zero Shot paradigm, research on LLM-based agents become a hot topic. Recent research on LLM-based agents, covering various resoning paradigms\cite{json2022cot,shun2023ReAct,shun2023tot, Maci2024got} and structures \cite{Char2023Mem, timo2023tool, Noah2023Reflexion}, has significantly improved single-agent capabilities. This has led to important advances across multiple fields \cite{hong2024data, autogpt_2024, wang2023voyager}. Inspired by the concept of swarm intelligence, the LLM-based MultiAgent System has become a hot subtopic in this research field. By integrating agents with different skills and characteristics into one system, this system can achieve more complex and advanced objectives than single agents. For example, in the fields of software engineering \cite{hong2023metagpt, qian2023communicative}, social simulation \cite{park2023generative, wang2023humanoid}, and building general problem-solving frameworks \cite{li2023camel, wu2023autogen, li2023modelscope}, multi-agent systems have demonstrated their ability to solve higher-level and broader range of problems.

With the development of VLM, the Agent has gradually shifted from a single text module to multi-modal interaction, and has derived fields that can interact with the real world, such as the Device Operation Agent(DOA)\cite{wang2024mobileagent,he2024webvoyager,zhang2023appagent}, robot control \cite{li2023vision}. However, in the field where VLM is used as the core of the Agent, research on using swarm intelligences to accomplish more complex tasks is still relatively less. We hope to supplement research in this field through MobileExperts, promoting the development of Agents in multi-modal multi-agent systems.

\subsection{Device Operation Agent}

Device Operation Agent (DOA) is an agent system that enables automation of human computing devices by configuring their action spaces, using either LLM or VLM. Currently, DOA mainly applies to three types of devices: web browsers, computers, and mobile devices. The present research trend focuses on developing User Interface (UI)-based agents that can automate user tasks on these devices. The progression of this type of research can be summarized via two primary directions.

First, UI-based Agents \cite{li2024personal} simulate basic user operations thereby transforming user tasks into a series of interactions with the device's UI. In this process, the core model of DOA is undergoing transformation.Earlier studies utilized the LLM to describe the sequence of actions for tasks by structuring device UI elements into textual data, such as HTML and XML formats. This LLM-centric approach was reflected across web, computer and mobile devices\cite{wen2024autodroid, zhou2023webarena}. However, with the development of VLM, end-to-end methods show greater potential as the core model of DOA and achieved superior performance in multiple platforms\cite{wu2024oscopilot, zhang2024ufo, he2024webvoyager, tan2024towards, wang2024mobileagent, zhang2023appagent}. But an increase in agents that apply VLM has introduced new challenges for DOA's development: accurately interpreting the meaning behind each UI element’s operation becomes significantly difficult for the universal multimodal models. This leads to the second direction in the development of DOA — enhancing the model's understanding of the UI. For instance, research like \cite{fuyu-8b, hong2023cogagent, niu2024screenagent, you2024ferretui, baechler2024screenai} carried out in-depth training on the UI elements of Android, Apple as well as PC operating systems, achieving comparable performance with large-scale parameter models with a smaller scale parameter volume.

As previously mentioned, UI-based Agents need to frequently call the model when executing operations, leading to costs in reasoning and time. For devices based on web browsers or computers, this problem can be solved by generating executable scripts with LLM to form tools\cite{zhang2024ufo, wu2024oscopilot}. However, due to the limitations of mobile device operating systems, this tool formation strategy is not applicable on mobile devices, impacting the operation efficiency of DOA on mobile devices. 

In response to this limitation, we propose the MobileExperts, aiming to develop an innovative tool formation strategy for mobile devices, effectively solving the above problems, reducing reasoning and time costs in a mobile environment.

\section{Methodology}

\begin{figure}[htbp]
  \centering
  \includegraphics[scale = 0.33]{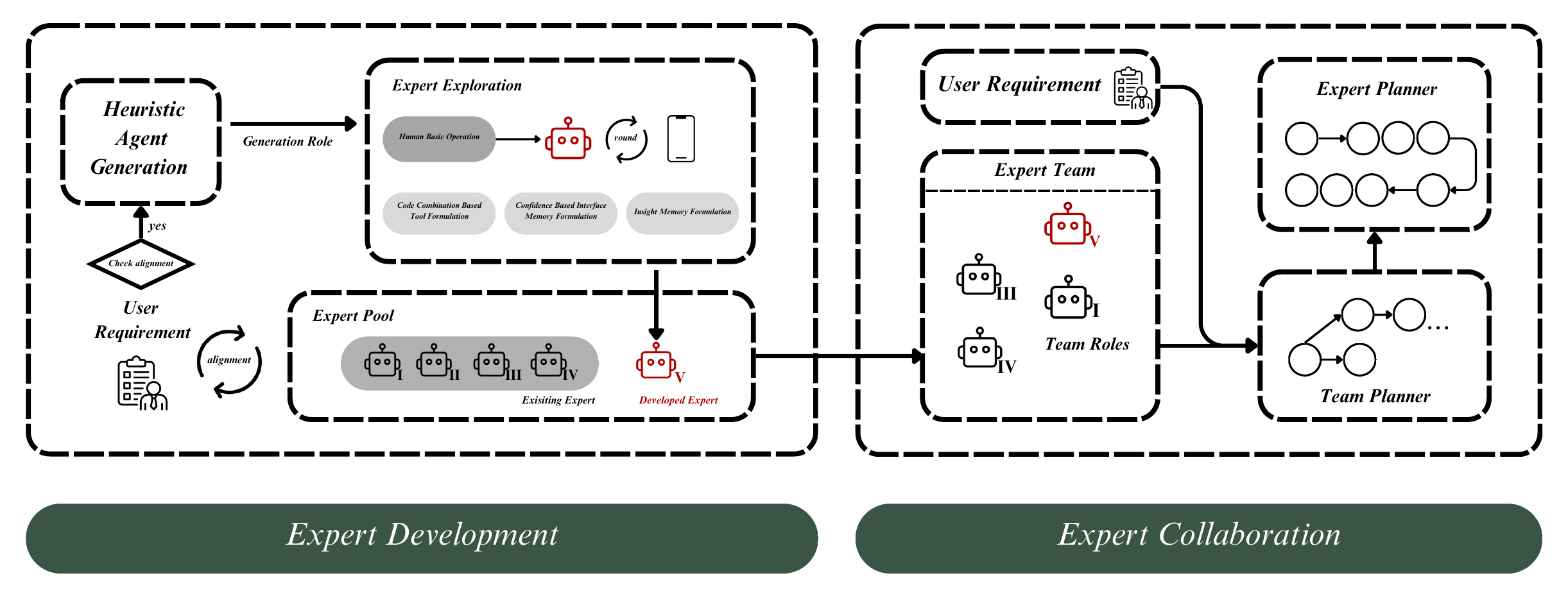}
  \captionsetup{font={small}}
  \caption{\textbf{The Framework of MobileExperts: }Upon fielding a user's request, the system undertakes an examination of the expert pool to identify an appropriate alignment to assemble the team. When no suitable expert is present, the system utilizes a heuristic algorithm to create a new expert. Subsequently, The newly minted experts then engage in the exploration process, accumulating experience to become competent for the task. After all the experts are in position, the execution process based on the expert's portraits generates a dual-layer plan, and then the experts complete the task in a self-verifying manner.}
  \label{framework of MobileExperts}
\end{figure}

In this section, we initially present an overview of MobileExperts, our proposed multi-agent collaborative framework for solving complex tasks in the field of mobile device operation agent, as shown in Figure \ref{framework of MobileExperts}. 

\subsection{Expert Development}

\begin{figure}[htbp] 
  \centering
  \includegraphics[scale = 0.33]{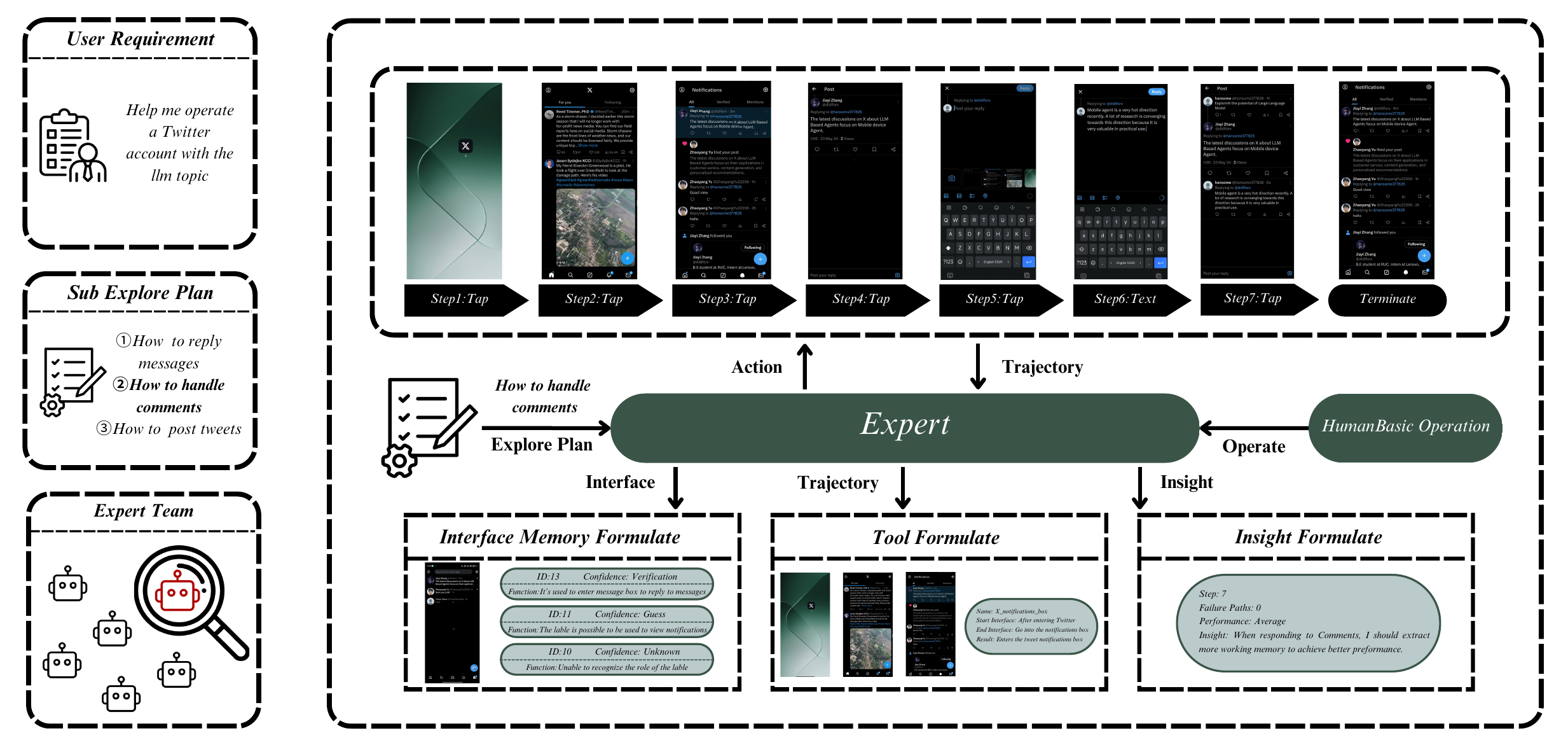}
  \captionsetup{font={small}}
  \caption{\textbf{Expert Exploration}: In the exploration phase, each agent receives tailored tasks, which are then broken down into sub-tasks to streamline the exploration process. Upon completion of a sub-task, whether successful or not, the agent extracts three types of memories from its trajectory: interface memories, procedural memories (tools), and insight memories for use in subsequent execution phases. It's a pivotal step for expert's development.}
  \label{framework of exploration}
\end{figure}

Ensuring that mobile device operating agents have high efficiency, strong generalization, and effective error correction across various devices is crucial for their successful integration into real-world applications. However, due to inherent differences among mobile devices, an agent not specifically trained for each device may struggle to meet these criteria. To address this, we've introduced an exploration stage designed to equip the agent with unique tools and knowledge, enabling it to become proficient on a specific mobile device. The exploration process of MobileExperts is illustrated in the figure \ref{framework of exploration}.

\subsubsection{Exploration Process}
Expert exploration is a pivotal phase in the mobileexperts. During this phase, each Expert $\mathcal{E}$ leverages their individual portraits to formulate multiple explore subtasks $\mathbb{S}$ aligned with the current requirement $\mathcal{R}$ as Equation \ref{eq:subtasks}

\begin{equation}\label{eq:subtasks}
    \mathbb{S} = \{r_1, r_2, \dotsm r_n\} = Decompose(\mathcal{E},\mathcal{R}),
\end{equation}

As the Expert engages in individual exploration phases, they tailor the generation of a tool set $\mathcal{T}$ and memories unique to user's device, including device-specific interface memories $\mathcal{M}_{\text{interface}}$ and insight memories $\mathcal{M}_{\text{insight}}$ as shown in Equation \ref{eq:single_agent_exploration_phase}. This process enables the Expert to become a domain specialist on the user's device, effectively addressing device-specific challenges. 

\begin{equation}\label{eq:single_agent_exploration_phase}
\mathcal{T}_i, \mathcal{M}_i^{\text{interface}}, \mathcal{M}_i^{\text{insight}} = \mathcal{E}(r_i)
\end{equation}


\subsubsection{Code Combination Based Tool Formulation}

In the mobile device environment, the mobile device operation agent faces complex challenges of learning different application usage methods. Current studies primarily concentrate on enhancing the agent's recognition of icons to tackle the challenge, neglecting the method of forming tools through exploration. The absence of such tools forces these agents to rely on a Visual Language Model (VLM) for each action, which has led to significant room for improvement in the  reasoning costs of the mobile device operation agent. To surmount this challenge, we introduce an innovative approach, Code Combination Based Action Extension. Specifically, we have defined a basic operational space for experts, drawing upon human behaviors. We task experts with the exploration of the user device to identify fixed operational workflows. By harnessing the encoding capabilities, we effectively connect these basic operations to form a series of new extended actions, achieving tool generation in the field of mobile device operation agents. 


\textbf{HumanBasic Operation.} Emulating human behavior, we craft nine basic operations for the expert, known as HumanBasic Operations. These encompass basic smartphone interactions, such as: tapping (Tap), text entry (Text), swiping (Swipe), reverting (Back), navigating to the home screen (Home), and terminating an operation (Stop). These operations extend beyond facilitating basic interaction, they also emulate the processes of human cognition. The \textit{Think} operation models human cognitive processes concerning short-term objectives. The \textit{Read} operation facilitates processing extensive documents, while the \textit{Wait} function ensures adaptability to the performance variations of different devices. Detailed operation parameters and descriptions are presented in the table \ref{humanbasic operation}.

\begin{table}[h]
\captionsetup{font={small}}
\caption{\textbf{HumanBasic Operation}'s parameters and descriptions}
\centering
\scalebox{0.65}{
\small
\begin{tabularx}{1.53\textwidth}{@{}lp{2cm}X@{}}
\toprule
\textbf{Operation }& \textbf{Params}                                         & \textbf{Description}                               \\ \midrule
Tap       & x, y                           & Simulates the human touch by selecting an element on the screen, used for choosing menu options or activating features. \\ \midrule
Text      & content                               & Inputs a specified string into a text field on the interface, used for filling out forms or searching for information.
                                           \\ \midrule
Swipe     & x, y, direction               & Navigates through content or performs actions by swiping in a particular direction on the screen, such as switching pages.                                           \\ \midrule
Read      & file, goal       &Reads the content of a specified document, such as PDFs or Word files, used for obtaining and analyzing information.                                           \\ \midrule
think     & flow, goal       &Used during task execution to read from or write to working memory, assisting in decision-making and planning.                  \\ \midrule 

Back      & none                                           &Returns to the previous interface, used for undoing actions or navigating to the previous screen.                                           \\ \midrule
Home      & none                                           &Navigates back to the smartphone's main screen, typically used to end the current session or access other applications.                                           \\ \midrule
Wait      & none                                           &Chooses to wait when the current operation is unresponsive or loading, ensuring the system has adequate time to process.
                                           \\ \midrule
Stop      & none                                           &Ends the exploration of the current task when it is believed to be completed or no longer feasible.                                           \\ \bottomrule
\end{tabularx}
}\label{humanbasic operation}
\end{table}

\textbf{WorkFlow Operation.} The absence of tool formulation in the field of mobile device operation agent is closely tied to the unique attributes of mobile devices. Contrasting with the proficiency of PC-based agents in creating and executing scripts for broad application control \cite{wang2023voyager, wu2024oscopilot}, mobile platform agents encounter limitations imposed by their APIs. These constraints, while hindering traditional script-based tool construction, also inspire us to conceive a new formation method for tools: code combination based tool formulation.

\begin{figure}[htbp]
  \centering
  \includegraphics[scale = 0.3]{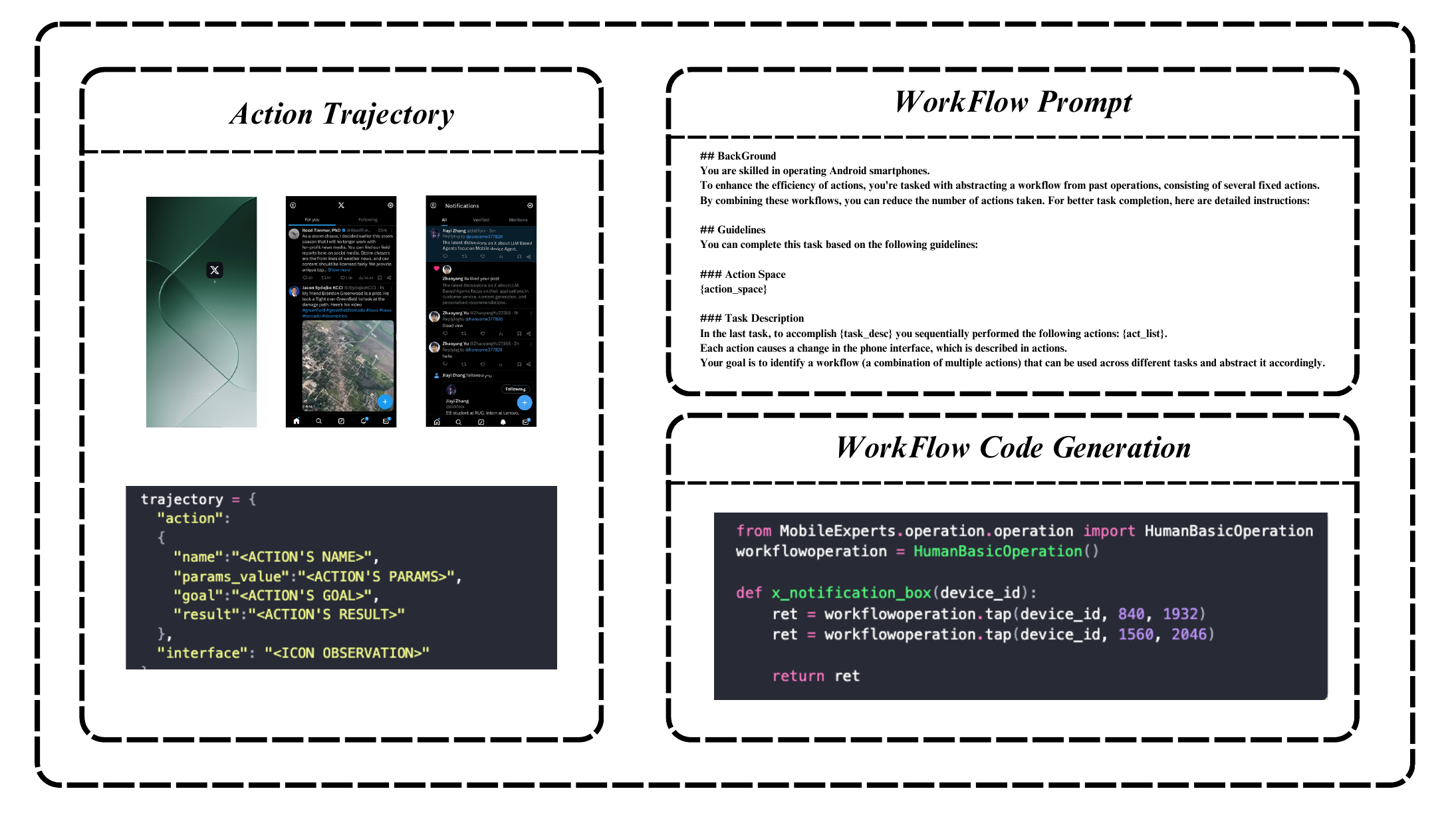}
  \captionsetup{font={small}}
  \caption{\textbf{Tool Formulation: }In the exploration phase, we documented action trajectories with names, parameters, goals, and results. Using workflow prompts, we recognized sequences of actions from fixed UI elements and make agent write tool code based on foundational classes, to achieve  automated tool formulation.}
  \label{tool formulation}
\end{figure}

In the applications of mobile devices, there are variable and immutable UI elements, and essentially, the interaction of the mobile device operation agent with mobile devices is the operation of these UI elements. By recording the action trajectory of the agent and the user interface in a specific exploration process of an app, the agent can identify a reusable workflow around multiple immutable positional elements. To enhance the efficiency and precision of workflow generation, the expert initiates by creating a rigorous description of the workflow, which encompass three key elements: the initial UI state, the final state, and a functional summary. Following that, by integrating the parameters from prior action trajectory and the code of each subsequent operation within humanbasic operation space, the expert can finalize the corresponding workflow operation function, thereby giving rise to a tool. In figure \ref{tool formulation}, we present an illustrative example of tool formulation using twitter's operational trajectory.


\subsubsection{Memory Formulation}
Accumulating memory from the Action Trajectory of exploration is a critical step to ensure the Expert has the ability to maintain its generalization capability and self-correction ability through device exploration. During the exploration stage, we pay special attention to the accumulation of memory of the functions of the user's interface icons, as well as the accumulation of insights closely related to the expert's own responsibilities.

\textbf{InterFace Memory} InterFace memory outlines a memory mechanism exclusively designed for device-specific icons. Due to the diversity of mobile device icon layouts and the widespread existence of system themes, generic VLMs struggle to directly discern the functions of different icons on the mobile interface. 

To address this issue, we have proposed a Confidence-based Icon Memory Mechanism. Differing from previous research\cite{zhang2023appagent}, our method encourages the expert to guess and verify the functions of the icons related to the task goal, instead of being limited to the icons that have been manipulated, achieving three types (\textbf{Verified, Hypothesized, Uncharted}) of confidence labelling for different icons. By forming such memories, we managed to enhance the scope of interface memory while maintaining the effects of interface exploration prompting more icons that might be related to tasks, and thereby minimizing the possibility that visual language models ignore icons that may be related to tasks due to hallucination.
 
\textbf{Insight Memory } One key distinction between experts and novices lies in their profound understanding of their responsibilities. Leveraging their insight into relevant tasks, experts can achieve superior performance with fewer steps. This insight is derived from reflecting on action trajectories. To this end, we have designed an insight memory mechanism that focuses on \textbf{task execution efficiency}, \textbf{failure paths}, and \textbf{performance}. By comprehensively considering these three elements, the agent can gain deep insights into the current exploration task, thereby achieving a leap from novice to expert in terms of responsibility fulfillment.

\subsection{Expert Collaboration}

Device Operation Agents face another crucial challenge in real-world applications: maintaining action consistency during high-frequency environmental interactions while effectively decomposing complex long-term tasks. To address this, we propose an innovative Expert Act Manner, ensuring consistency and efficiency in the Agent's execution of tasks. Furthermore, inspired by \cite{hong2024data}, we introduce a dual-layer planning mechanism at both Team and Expert levels, transforming complex tasks into atomic tasks with shorter execution paths.

\begin{figure}[htbp]
  \centering
  \includegraphics[scale = 0.35]{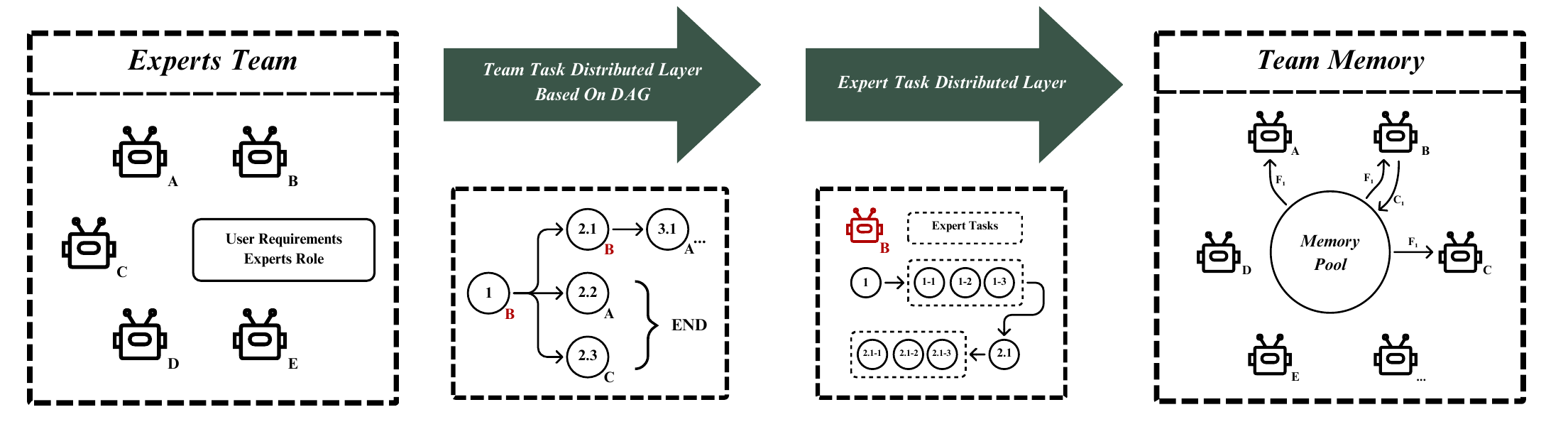}
  \captionsetup{font={small}}
  \caption{\textbf{Collaboration mechanism}: MobileExperts employs Double Layer Planning to decompose complex tasks and construct Team Memory. Each Expert selectively interacts with the Team Memory through Commit and Fetch methods, enabling effective collaboration.}
  
  \label{dual_planner}
\end{figure}

\subsubsection{Expert Act Manner}

The fundamental principle behind maintaining action consistency in Device Operation Agents lies in its role as the foundation for agent decision-making.  In previous studies, experts typically employed the React paradigm to reason about and interact with the current environment, and enhancing their perception of the current state by summarizing action trajectories. While this approach maintains consistency in atomic tasks, it often results in significant information loss when confronting long-term tasks, consequently compromising the decision accuracy of Operation Agents. To mitigate this limitation, we propose an enhanced agent behavior paradigm coupled with a proactive working memory processing function.

\textbf{Self Verifying Manner: }We propose self verifying manner, which enhances expert's ability on visual-linguistic tasks by incorporating dual-state visual information input and an 'Observe-Verify-Act' cycle. This mechanism continuously comparing previous and current visual states while considering action objectives. This approach enables the expert to maintain better consistency between consecutive actions, resulting in more precise and context-aware decision-making in device operation tasks.

\textbf{Proactive Working Memory: }During Expert Level's task execution, critical information generated by the Expert is stored as its Working Memory. For long-term tasks, simply placing accumulated extensive Working Memory within the prompt leads to increased token and time costs, while excess information may also affect the Expert's decision-making ability. To address this issue, we've incorporated a Proactive Working Memory Processing function as a BasicOperation, allowing the Expert to efficiently manage its working memory during task execution. This ensures the Expert maintains action consistency during long-term tasks while reducing token costs.

\subsubsection{Expert Collaboration Via Double-Layer Planning}

Accomplishing complex and long-term tasks, such as managing a Twitter account, presents two major challenges for Experts. The first challenge lies in decomposing the task into atomic task that an Expert can complete. The second challenge involves executing these tasks with high quality.  In real-world scenarios, complex tasks typically require team collaboration. Therefore, developing an appropriate collaborative mechanism for Experts becomes crucial in addressing this issue. We tackle these challenges by implementing a double-layer planning mechanism and a team-shared memory pool, enabling effective collaboration among Experts.

Through Expert Development, Experts with diverse capabilities and roles form a team. Double layer planning operates at both team and expert levels to plan user instructions. Specifically, the team task distributed Layer decomposes instructions into a graph organized as a DAG (Directed Acyclic Graph) based on user directives and team member role descriptions. Each node in this graph represents a task assigned to an Expert with corresponding abilities.

This graph structure also establishes a communication mechanism among Experts. By maintaining a team-shared Memory Pool, each Expert uses the Commit method to store structured descriptions of task states and accumulated working memories upon completion of Team Level tasks. Experts dependent on these tasks then use the Fetch method to selectively extract relevant information, storing it in their individual working memories to support current task execution.

At the expert task decomposition layer, the planner breaks down Team Level task descriptions into sequences of atomic tasks. Based on the status of these atomic tasks, the Planner can dynamically adjust the sequence order or repeat execution, implementing a flexible regulation mechanism to ensure successful completion of atomic tasks.




\section{Experiments}

In order to comprehensively evaluate the ability of MobileExperts in handling diversified and complex tasks, we have constructed a new benchmark - Expert-eval. Based on this Benchmark, we compared our method with two baselines, and performed ablation experiments on various key modules of MobileExperts. Meanwhile, in order to intuitively display the performance of MobileExperts in tasks of different intelligence levels, we have carried out a case study to deeply analyze its behavior strategy and effect. Considering the requirement of overall action space for api accessibility, we chose to conduct these experiments on the Android system. The overall tasks included in Expert-Eval can be found in Table \ref{expert-eval}.


\begin{table}[h]
\captionsetup{font={small}}
\caption{Expert Eval: We developed 28 examples encompassing various complexity levels and application scenarios to comprehensively assess the performance of Device Operation Agent in practical applications. }
\centering
\scalebox{0.66}{
\small
\begin{tabularx}{1.5\textwidth}{@{}cccX@{}}
\toprule
\textbf{Category}                           & \textbf{Application}           & \textbf{Complexity} & \textbf{Task Description}                                                                         \\ \midrule
\multirow{8}{*}{\textbf{Social Media}}      & \multirow{2}{*}{Twitter}             & C1                  & Open the messages on twitter and read the first unread message.                                               \\ \cmidrule(l){3-4} 
                                            &                                & C2                  & Post a tweet summarizing the recent discussions on twitter about LLM Based Agents.                      \\ \cmidrule(l){2-4} 
                                            & \multirow{2}{*}{Instagram}     & C1                  & Search for "lenovo" on Instagram.                                                                 \\ \cmidrule(l){3-4} 
                                            &                                & C2                  & Search for meme-related accounts on Instagram and post comments based on the content.             \\ \cmidrule(l){2-4} 
                                            & \multirow{2}{*}{Tiktok}        & C1                  & Leave a comment on tiktok's latest video.                                                         \\ \cmidrule(l){3-4} 
                                            &                                & C2                  & Review the last ten videos on my TikTok feed trends and search for topics that might interest me. \\ \cmidrule(l){2-4} 
                                            & \multirow{2}{*}{Xiaohongshu}   & C1                  & Scroll through Xiaohongshu until you find a post that explains a trip to Miami, and then like it. \\ \cmidrule(l){3-4} 
                                            &                                & C2                  & Find popular food in Miami and then create a collection album in Xiaohongshu.                      \\ \midrule
\multirow{8}{*}{\textbf{Online Service}}    & \multirow{2}{*}{Google Maps}   & C1                  & Navigate to Renmin University of China.                                                         \\ \cmidrule(l){3-4} 
                                            &                                & C2                  & Find delicious Xinjiang cuisine near Renmin University of China and then navigate there.          \\ \cmidrule(l){2-4} 
                                            & \multirow{2}{*}{Alibaba}       & C1                  & Enter Lenovo's store in Temu.                                                                      \\ \cmidrule(l){3-4} 
                                            &                                & C2                  & Find the price of the latest Lenovo Legion model.                                                  \\ \cmidrule(l){2-4} 
                                            & \multirow{2}{*}{Google Play}   & C1                  & Download the JD (Jingdong) App from Google Play.                                                   \\ \cmidrule(l){3-4} 
                                            &                                & C2                  & Download a popular note-taking software.                                                           \\ \cmidrule(l){2-4} 
                                            & \multirow{2}{*}{Messenger}     & C1                  & Check the latest verification code information.                                                    \\ \cmidrule(l){3-4} 
                                            &                                & C2                  & Delete messages in the text that are unrelated to the verification code.                \\ \midrule
\multirow{8}{*}{\textbf{Productivity Tool}} & \multirow{2}{*}{Notion}        & C1                  & Create a new page in Notion.                                                                       \\ \cmidrule(l){3-4} 
                                            &                                & C2                  & Create a new note detailing your job duties and title it appropriately.                     \\ \cmidrule(l){2-4} 
                                            & \multirow{2}{*}{Gmail}         & C1                  & Write an email to \{address\}.                                                                  \\ \cmidrule(l){3-4} 
                                            &                                & C2                  & Go through the unread emails and draft a response based on their content.                        \\ \cmidrule(l){2-4} 
                                            & \multirow{2}{*}{Google Chrome} & C1                  & Enter arxiv.org in Chrome.                                                                               \\ \cmidrule(l){3-4} 
                                            &                                & C2                  & Download the latest paper on LLM (Large Language Model) Based Agent from arXiv.                    \\ \cmidrule(l){2-4} 
                                            & \multirow{2}{*}{Wikipedia}     & C1                  & Open Wikipedia and search for "collective intelligence".                                           \\ \cmidrule(l){3-4} 
                                            &                                & C2                  & Create a Wikipedia page about yourself.                                                   \\ \midrule
\multicolumn{2}{c}{\multirow{6}{*}{\textbf{Cross-App}}}                      & C2                  & Download \{app\_name\} and complete account registration with my phone number.                    \\ \cmidrule(l){3-4} 
\multicolumn{2}{c}{}                                                         & C3                  & Help me operate a twitter account with the theme \{topic\}.                                            \\ \cmidrule(l){3-4} 
\multicolumn{2}{c}{}                                                         & C3                  & Help me find \{Product Name\} that is popular and reasonably priced.                              \\ \cmidrule(l){3-4} 
\multicolumn{2}{c}{}                                                         & C3                  & Help me explore how to use \{app\_name\} and write a technical report-style document in Notion.   \\ \bottomrule
\end{tabularx}
}
\label{expert-eval}
\end{table}

\subsection{Settings}
\textbf{Expert-Eval.} In order to comprehensively evaluate MobileExperts' performance in handling diverse applications and tasks of different complexities, we have designed a new benchmark named Expert-Eval.
Expert-Eval aims to measure the intelligence level of mobile device operation agent at varying levels, which is closely related to the planning capabilities required for the task. We have defined three intelligence levels, ranging from simple to complex as follows:

\begin{itemize}[leftmargin=2em]
\setlength{\itemsep}{1pt}
\setlength{\parsep}{1pt}
\setlength{\parskip}{1pt}
    \item C1 (Executor): Task execution is directly in accordance with the user's instructions, with no additional planning required.
    \item C2 (Planner): The task requires basic planning, the system needs to autonomously plan and execute tasks with clear objectives.
    \item C3 (Strategist): The task requires complex, long-term planning, the system needs to formulate detailed plans for tasks of high complexity with clear objectives, and make adjustments during execution based on feedback.
\end{itemize}

In addition, in order to verify the generalization ability of MobileExperts in different application scenarios, we divide applications into three main categories: Social Media, Online Service, and Productivity Tool. Based on these classifications, we have designed a series of tasks that require the use of one or more applications to test the adaptability and flexibility of MobileExperts in actual applications.

\textbf{Metrics.} We design five metrics to measure the performance of the mobile device operation agent in various dimensions.
\begin{itemize}[leftmargin=2em]
\setlength{\itemsep}{1pt}
\setlength{\parsep}{1pt}
\setlength{\parskip}{1pt}
    \item \textbf{Success Rate}: This metric measures whether the agent successfully completes tasks. If the task is completed, the score is marked as 1.
    \item \textbf{Process Score}: This metric assesses the action score of the agent during the execution of the task. When facing tasks that require complex planning ability, even if the agent cannot complete all actions, its effective actions will have a positive impact on this metric.
    \item \textbf{Reasoning Steps}: This metric measures the efficiency of the agent in executing tasks. For mobile device operation agents, the main time consumption is to choose actions through calling the VLM. This metric is equal to the number of times the agent calls the VLM during the execution of the task.
    \item \textbf{Complete Performance}: Unlike the performance score, this metric measures the quality of the agent's task completion. For C1 level instruction obedience tasks, it can get full marks (10 points) as long as it is completed. For tasks of C2 and C3 levels, there is a significant difference in the effect of completing tasks by different methods. We use the visual language model to score the initial goal of the task and the action track of the agent completing the task. 
\end{itemize}

\subsection{Experiment Results}

In this experiment, we used a unified mobile device, pre-installed with all the necessary applications, and tested expert-eval on three methods. In order to ensure the consistency of the experimental results, we set the maximum number of automatic explorations for individual tasks of MobileExperts and AppAgent to 10 times, and required each method to try a maximum of three times, finally choosing the best results as the outcome. In addition, considering the importance of the process score, we set the maximum number of execution steps for all methods's C1 and C2 Task to 15 steps to prevent cycle errors, ensuring the efficiency of the experiment and the reliability of the results. The results of the experiment are presented in table \ref{exp_result}.




\begin{table}[h]
\captionsetup{font={small}}
\caption{Performance of MobileExperts, AppAgent\cite{zhang2023appagent} and MobileAgent\cite{wang2024mobileagent} on Expert Eval.}
\centering
\scalebox{0.75}{
\small
\begin{tabular}{@{}c|llllllllllll@{}}
\toprule
\multirow{2}{*}{Method} &
  \multicolumn{4}{c}{Complexity 1} &
  \multicolumn{4}{c}{Complexity 2} &
  \multicolumn{4}{c}{Complexity 3} \\ \cmidrule(l){2-13} 
 &
  \multicolumn{1}{c}{SU} &
  \multicolumn{1}{c}{PS} &
  \multicolumn{1}{c}{RS} &
  \multicolumn{1}{c|}{CP} &
  \multicolumn{1}{c}{SU} &
  \multicolumn{1}{c}{PS} &
  \multicolumn{1}{c}{RS} &
  \multicolumn{1}{c|}{CP} &
  \multicolumn{1}{c}{SU} &
  \multicolumn{1}{c}{PS} &
  \multicolumn{1}{c}{RS} &
  \multicolumn{1}{c}{CP} \\ \midrule
Mobile Experts &
  \textbf{0.8333} &
  \textbf{0.7605} &
  \textbf{5.9167} &
  \multicolumn{1}{l|}{\textbf{8.3333}} &
  \textbf{0.7692} &
  \textbf{0.6739} &
  \textbf{7.1538} &
  \multicolumn{1}{l|}{\textbf{7.3077}} &
  \textbf{1.0000} &
  \textbf{0.5408} &
  \textbf{28.3333} &
  \textbf{7.0000} \\
App Agent &
  0.7500 &
  0.7345 &
  7.4167 &
  \multicolumn{1}{l|}{7.5000} &
  0.6154 &
  0.6174 &
  8.9231 &
  \multicolumn{1}{l|}{5.2308} &
  0.3333 &
  0.3754 &
  33.6667 &
  2.6667 \\
Mobile Agent &
  0.7500 &
  0.7537 &
  6.0000 &
  \multicolumn{1}{l|}{7.5000} &
  0.2308 &
  0.5095 &
  9.3077 &
  \multicolumn{1}{l|}{1.9231} &
  0.3333 &
  0.4014 &
  38.3333 &
  2.3333 \\ \bottomrule
\end{tabular}
}
\label{exp_result}
\end{table}



\section{Conclusion}
This paper introduces MobileExperts, an innovative multimodal agent framework designed to address complex mobile device operation tasks. The framework achieves breakthrough performance through two core features: code combination based tool formulation on mobile devices and expert collaboration via double-layer planning. This approach not only provides a novel method for automatic tool formulation in the field of mobile device operation agents but also introduces innovative action strategies for agent teams. Experimental results demonstrate that MobileExperts excels in handling high-complexity tasks while significantly reducing reasoning costs. These findings not only confirm the framework's potential to enhance agent capabilities and reduce resources consumption but also advance research in mobile device operation automation.

\newpage







\end{document}